\pdfoutput=1
\documentclass[letterpaper,10pt,journal]{IEEEtran}
\usepackage[T1]{fontenc}
\usepackage[protrusion=true,expansion=false,final]{microtype}
\usepackage{amsmath,amssymb,amsfonts}
\usepackage[varg]{newtxmath}
\usepackage{graphicx}
\usepackage{booktabs}
\usepackage{balance}
\usepackage{url}
\usepackage{xcolor}
\usepackage[colorlinks=true,linkcolor=blue,citecolor=blue,urlcolor=blue,
            filecolor=blue]{hyperref}
\providecommand{\tightlist}{\setlength{\itemsep}{0pt}\setlength{\parskip}{0pt}}

\makeatletter
\def\section{\@startsection{section}{1}{\z@}{2.2ex plus 1ex minus .5ex}%
{0.5ex plus .3ex}{\normalfont\normalsize\centering\scshape}}%
\def\subsection{\@startsection{subsection}{2}{\z@}{2.4ex plus 1ex minus .5ex}%
{0.5ex plus .3ex}{\normalfont\normalsize\itshape}}%
\makeatother
\begin{document}
\title{DART-S: Reachability-Audited Active-Suspension Preconditioning
for Off-Road Vehicle Jumps}
\author{Yu Hu\textsuperscript{1,2,*}, Fangzhou Zhao\textsuperscript{1},
Liang Chen\textsuperscript{1}, Chen Min\textsuperscript{1},
Wei Li\textsuperscript{1,2}, Mingyuan Sang\textsuperscript{1,2},\\
Jiajia Ma\textsuperscript{3}, Shican Chen\textsuperscript{3},
Di Pang\textsuperscript{3}, Baolei Chen\textsuperscript{3}
\thanks{\textsuperscript{1}Research Center for Intelligent Computing Systems,
Institute of Computing Technology, CAS, Beijing 100190, China.}
\thanks{\textsuperscript{2}School of Computer Science and Technology,
University of Chinese Academy of Sciences, Beijing 100049, China.}
\thanks{\textsuperscript{3}Dong Feng Off-Road Vehicle Co., Ltd,
Shiyan 442000, China.}
\thanks{\textsuperscript{*}Corresponding author: Yu Hu
(e-mail: \href{mailto:huyu@ict.ac.cn}{huyu@ict.ac.cn}).}}
\maketitle
\begin{abstract}
Airborne torque reaction cannot recover takeoff errors beyond the wheel
angular-momentum budget. DART-S applies ramp-face suspension preconditioning
to change pitch, pitch rate, and wheel spin before liftoff, thereby shifting
the queried state and altering the remaining authority budget. To predict how
each suspension action reshapes this state--budget pair, DART-S employs a local
calibration map. A
support-aware selector combines the predicted shift with local outcome
evidence and an interval-reachability screen; an exact-pair audit reports
residual authority. Across 600 new runs in 72 independent BeamNG sessions,
every positive, negative, and boundary query follows its prespecified branch.
At the confirmed 40$^\circ$/13~m/s boundary, DART-S attains 24/24 post-touchdown
attitude-criterion successes versus 0/24 for DART (session-level
Holm-adjusted \(p=0.0234\)). At 11.5~m/s, a 0.35~s timing action attains
23/24 versus 0/24 for the static preset (\(p=0.0156\)). The 200~rad/s
command guard keeps drivetrain hard-limit exceedance at zero across all 600
runs.
The source code will be available at \url{https://github.com/MeridianCAS/DART-S}.
\end{abstract}
\section{Introduction}

Airborne torque reaction cannot recover an unfavorable takeoff state when
the required correction exceeds the available wheel angular momentum. In
our prior work, DART \cite{r1} quantified this boundary on a platform with a mass of 1383~kg:
the recoverable pitch-rate budget is 9--13~deg/s in the tighter nose-up
direction, about twice that in the reverse-inclusive braking direction, and
only 16--18~deg/s nose-up at the drivetrain hard limit. Wheel angular momentum,
not motor torque, bounds the available correction. DART back-propagates the
landing constraint into a feasible-takeoff set and can refuse a jump or reshape
takeoff speed, but it lacks an independent ground-phase actuator for direct
shaping of takeoff pitch and pitch rate before liftoff.

DART-S moves this control leverage before liftoff by repurposing per-corner
active suspension as a takeoff-state actuator. Its core departure from
attitude-only preconditioning is two-sided: the ramp-face preset changes
takeoff pitch, takeoff pitch rate, and individual-wheel spin, so it moves the state queried by the
reachability test while changing the angular-momentum budget that defines
the tested set. The ground phase thus creates the initial condition and
remaining authority from which the airborne phase completes the correction.
Motocross riders have long preloaded suspension on the ramp face to alter
launch trajectory \cite{r2}. DART-S formalizes this technique through full-size
automation and a feasibility audit.

Related actuator technologies have been announced for BYD production vehicles:
the DiSus portfolio comprises intelligent damping, air-body-control, and
hydraulic-body-control branches \cite{r3}. Existing patented functions employ
suspension or wheel torque for in-flight stabilization, landing preparation,
or driveline protection \cite{r4}, \cite{r5}, \cite{r6}, rather than for pre-takeoff
state-and-budget shaping. To our knowledge, DART-S is the first framework to
map a timed pre-takeoff suspension action to both the takeoff state and
remaining wheel-momentum budget, then compose it with budget-limited airborne
control.
We evaluate this coupling in deterministic full-scale soft-body simulation.
The empirical selector determines whether preconditioning is applied, while
the exact-pair certificate quantifies residual airborne authority.

\textbf{DART-S} adds a suspension-preconditioning channel before DART's
speed gate and airborne controller, as shown in Fig.~1. Our
contributions are:

\begin{figure*}[t]\centering
\includegraphics[width=\textwidth]{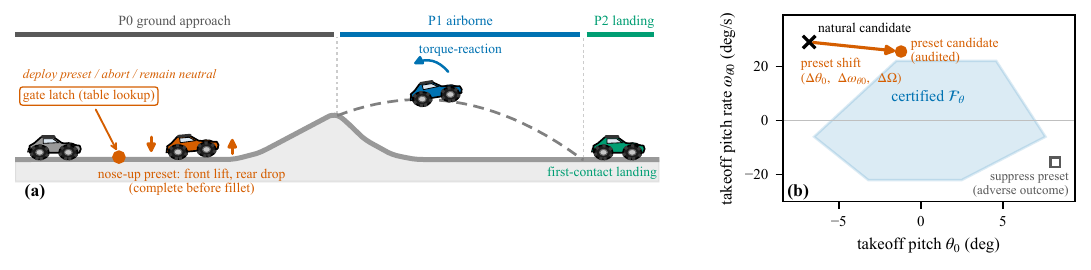}
\caption{DART-S overview. (a) Phase structure: during the approach (P0) the selector latches a per-condition decision (deploy the preset, abort, or remain neutral) early enough that the preset completes before the ramp fillet; DART's pitch-axis torque-reaction law corrects the airborne phase (P1) and landing metrics latch at first contact (P2). (b) Support-aware selector: calibrated attitude and wheel-spin shifts are evaluated by an operational interval-reachability screen. Local calibration support, outcome evidence, and that screen govern deployment; the exact-pair set reports residual authority diagnostically.}\label{fig:darts_fig1}
\end{figure*}

\begin{enumerate}
\def\labelenumi{\arabic{enumi}.}
\tightlist
\item
  \textbf{Two-sided actuation of a reachability query.} A per-corner ramp-face
  preset turns takeoff pitch and pitch rate from measured initial conditions
  into control variables while changing individual-wheel spin and hence the directional airborne
  budget. DART-S therefore acts on both the queried state and the set that
  evaluates it, rather than translating a state inside a fixed envelope.
\item
  \textbf{Timing-conditioned takeoff-state shaping under sign-changing suspension
  dynamics.} Ramp-face suspension does not produce a monotonic landing
  response: a favorable takeoff-angle shift can coincide with an adverse
  pitch-rate shift because release timing changes the contact transient.
  Independent five-speed evidence identifies the resulting 11.5--12~m/s
  release-phase valley.
  DART-S therefore treats release timing as a control variable and
  combines a geometry-conditioned action map with a curvature-saturation
  safeguard to select among no preset, a static preset, and a timed preset.
\item
  \textbf{Support-aware cross-phase composition.}
  We combine ground preconditioning, a spin-limited airborne law, and
  evidence-based abstention. Local outcome evidence determines deployment,
  while interval reachability screens the action-conditioned state and
  directional wheel budget; exact-pair reachability remains diagnostic.
  Independent BeamNG sessions confirm the boundary and timing results
  together with guarded branch fidelity.
\end{enumerate}

\section{Related Work}

Relevant prior work falls into three areas: airborne attitude control for
wheeled vehicles, active-suspension control around jumps, and pre-takeoff
state planning.

Momentum-exchange attitude work includes reaction-wheel jump-and-balance
platforms \cite{r7}, tail-assisted aerial systems \cite{r8}, \cite{r9}, and legged takeoff
shaping \cite{r10}, \cite{r11}, \cite{r12}. Wheel-reaction control is demonstrated in hardware \cite{r13}
and simulation \cite{r14}; learned in-air maneuvering relies on throttle/steering
with explicit wheel-speed and actuation limits \cite{r15}. A patent also uses
individual-wheel torque for jump stabilization \cite{r5}. None maps a pre-takeoff
suspension action to both launch state and landing-relevant wheel momentum.

Suspension work spans semi-active and production-oriented systems \cite{r16}, \cite{r17},
simulation-based EV vibration suppression \cite{r18}, attitude MPC with bump-road
tests \cite{r19}, preview-MPC obstacle crossing \cite{r20}, and full-scale hydraulic
sim-to-real transfer \cite{r21}. These regulate ground-contact motion. Jump-specific
systems react after airborne detection or protect driveline reconnection
\cite{r4}, \cite{r6}; the cited BYD release \cite{r3} documents no ramp-face preconditioning
or wheel-momentum coupling. This gap defines the pre-takeoff state-and-budget
interface addressed by DART-S.

Closest conceptually, motocross preload changes launch trajectory \cite{r2}, a
rallycross simulation finds that stiffer rear damping reduces pitch rate and
rear-axle kick-up \cite{r22}, and legged systems optimize takeoff and stance-force
profiles \cite{r10}, \cite{r11}, \cite{r12}. MPPI and traversability work addresses aggressive-driving
and terrain costs \cite{r23}, \cite{r24}, including when to jump \cite{r25}, but not a remaining
wheel-momentum feasibility audit. DART \cite{r1} supplies that audit and shapes
takeoff speed; DART-S instead couples direct pitch/rate shaping to wheel spin
and the resulting airborne budget.

Two closer precedents clarify the remaining novelty. A Jaguar Land Rover
patent changes relative leading/trailing-wheel ride height for ramp clearance
\cite{r26}; Jing \emph{et al.} optimize state and actuator trajectories using
contact-implicit trajectory optimization, validated in Webots \cite{r27}; and an
ICMA 2025 study designs an active suspension that launches a light off-road
vehicle and verifies feasibility in Amesim \cite{r28}. They establish pre-ramp
suspension actuation, but not analytic state-and-budget authorization coupled
to airborne control. DART-S contributes this state-and-budget interface,
rather than pre-ramp suspension actuation alone.

\section{Method}\label{sec:method}

DART-S acts on both the queried takeoff state and the wheel-spin-dependent
authority budget that defines its feasible set. This section first formulates
this two-sided reachability model, then describes suspension preconditioning,
deployment gating, and sequential cross-phase control.

\subsection{Phase structure and takeoff-state definition}

We inherit DART's three-phase structure: ground approach (P0), airborne (P1),
and landing (P2). DART-S defines the takeoff state
\(x_0=(v_0,\theta_0,\omega_{\theta0},\Omega_{1,0},\ldots,\Omega_{4,0})\), where \(v_0\) is ground-plane takeoff speed,
\(\theta_0\) is pitch angle, \(\omega_{\theta0}\) is the body-y pitch-rate component at the last ground
contact, and \(\Omega_{i,0}\) is the measured spin of wheel \(i\); pitch and pitch
rate are positive nose-up.
DART-S chooses a ground action \(a_s=(u,\tau)\), where \(u\) is the four-corner
reference-length command and \(\tau\) is either empty for a static preset or
the lip-relative schedule \((t_{\mathrm{lead}},T_{\mathrm{pulse}},
\gamma_{\mathrm{release}})\). Here \(t_{\mathrm{lead}}\) is the lead time before
predicted lip arrival, \(T_{\mathrm{pulse}}\) is the release duration, and
\(\gamma_{\mathrm{release}}\) is the fraction of the full preset released
toward neutral. DART-S treats this schedule as a deployable action. Measured
rebound phase is retained offline to evaluate phase-augmented takeoff
prediction. Deployment selects from neutral, static-preset, and confirmed
timed-preset actions without interpolating command amplitude or schedule.

DART's full landing constraint is \(\mathcal{C}_L\). We distinguish its airborne pitch
slice \(C_{\mathrm{air}}=\{|\theta-\theta_{\mathrm{surf}}|\le\bar\theta,\,
|\omega_\theta|\le\bar\omega\}\) from the experimental post-touchdown criterion
\(C_{\mathrm{post}}\), which also requires bounded roll, no tumble, and a
stable observation window. Here \(\theta_{\mathrm{surf}}\) is the
terrain-matched landing-surface pitch, and \(\mathcal{F}_\theta\) denotes the certified
preimage of \(C_{\mathrm{air}}\). At the
decision latch, the selector predicts a reference takeoff state
\(\hat{x}_0^{\mathrm{ref}}\) and forms each candidate
\(\hat{x}_0^c=\hat{x}_0^{\mathrm{ref}}+\Delta x_0^c\) from the calibrated
action-induced shift. For nominal auditing, we apply the calibrated mean-wheel
shift \(\Delta\bar\Omega^c\) to each predicted wheel and define
\(B_\uparrow^c=\sum_i I_i[\Omega_{\max}-\hat{\Omega}_{i,0}^c]_+/I_{yy}\) and
\(B_\downarrow^c=\sum_i I_i[\hat{\Omega}_{i,0}^c-\Omega_{\min}]_+/I_{yy}\),
where \([z]_+=\max(0,z)\). The mean-wheel shift supplies the nominal budget
audit, while local outcome evidence and the interval transition determine
deployment. We adopt the baseline flight-time estimate stored with the local
condition rather than predicting an action-conditioned \(\Delta T\).
After liftoff, a realized audit recomputes membership from the measured
\(x_0^c\). Thus candidate \(c\) changes both the queried pair and the geometry
of \(\mathcal{F}_\theta\); hats are omitted below when the same certificate applies to either
query.

For the exact terminal pair \((\theta_{\mathrm{surf}},0)\), DART's
null-then-correct certificate \cite{r1} first checks directional rate feasibility,
\(-B_\uparrow\le\omega_{\theta0}\le B_\downarrow\) and
\(|\omega_{\theta0}|/a_\theta\le T\), then checks the residual pitch displacement
against the closed-form budget-limited envelope of DART Theorem 3. DART-S
does not alter this certificate; it changes its state and budget inputs.
Consequently, \(F_\theta\) remains a sufficient inner approximation: failure
means uncertified, while a rate-bound or correction-time violation proves
the exact terminal rate unreachable. Nonzero
\((\bar\theta,\bar\omega)\) supplies an interval audit. Together, both forms
supply the selector's airborne audit.

\subsection{Active suspension as a takeoff-state actuator}

We repurpose the suspension as a slow takeoff boundary-condition actuator rather
than a landing absorber. We command the full preset before ramp loading.
A static action holds it through takeoff; a timing-conditioned action briefly
moves it toward neutral and then restores the identical final command before
liftoff. Each corner carries a series-hydraulic element that
scales the spring's reference length by \(1 + u_i \cdot \rho\), with per-corner command
\(u_i \in [-1, 1]\) and stroke ratio \(\rho\). The stroke time is long relative to the
torque loop, which motivates the early command. At \(u=0\) the element is
identical to the stock coilover. A \emph{nose-up} preset \(u=(+1,+1,-1,-1)\) (front
lift, rear drop) biases the takeoff state through two mechanisms:

\begin{enumerate}
\def\labelenumi{\arabic{enumi}.}
\tightlist
\item
  A \emph{static geometric term} raises the front contact patches by \(\Delta z_f\) and
  drops the rear by \(\Delta z_r\), pitching the body by
  \(\Delta\theta_{\mathrm{static}}\) $\approx$ arctan((\(\Delta z_f\) - \(\Delta z_r\))/L) with wheelbase L. Tire and bushing
  compliance absorb part of the commanded stroke, so the effective lever
  ratio is identified once per platform from static ground tests.
\item
  A \emph{lip-release dynamic term} arises as the front axle clears the lip:
  its preset stroke unloads while the rear still rides the ramp face,
  imparting a pitch-rate increment \(\Delta\omega_{\theta0}\). Its dependence on geometry and
  speed is represented by the calibration map rather than assumed
  analytically.
\end{enumerate}

For a timing-conditioned action, the controller estimates
\(t_{\mathrm{lip}}=d_{\mathrm{lip}}/v\). When
\(t_{\mathrm{lip}}\le t_{\mathrm{lead}}\), it commands
\((1-\gamma_{\mathrm{release}})u\) for \(T_{\mathrm{pulse}}\) and then restores
\(u\). The schedule
\((t_{\mathrm{lead}},T_{\mathrm{pulse}},\gamma_{\mathrm{release}})\) is
calibrated per operating condition; timing entries remain valid
only within their calibrated neighborhood.

Together these mechanisms define the cross-phase interface. We identify
them jointly from paired preset--neutral comparisons under matched commanded
entry conditions and fit a local empirical response map over geometry, decision-point
speed, and discrete action identity:

\(\operatorname{CalMap}(g,v,a_s)
  =(\Delta\theta_0,\Delta\omega_{\theta0},\Delta\bar\Omega)\).

\noindent Here \(g\) denotes the lip-geometry descriptor, including ramp angle
and lip radius.
Static entries set \(\tau=\varnothing\); timing entries require an
independent confirmatory evaluation before deployment. Within a declared
ramp-profile class and action identity, inverse-distance interpolation returns
the scalar shift and nearest support distance. The implementation requires
the query to remain inside the calibration points' axis-aligned support box
and distance threshold. Together they define the validated interpolation
envelope; queries outside it are routed to DART. Because the
deployed table never selects nose-down, the nose-down branch of the map
serves diagnosis only and is obtained by mirroring the nose-up response
rather than by separate calibration.
CalMap records the \emph{net
takeoff-state} change \((\Delta\theta_0,\Delta\omega_{\theta0},\Delta\bar\Omega)\)
rather than a landing prediction:
its attitude components shift the queried state, and its mean-wheel
component supplies a nominal budget audit. At liftoff, the realized audit
replaces that approximation with measured per-wheel takeoff speeds. Attitude
increments may be propagated via \(\Delta_{\mathrm{land}}\approx\Delta\theta_0+\Delta\omega_{\theta0}T\) for interpretation.
Deployment relies on the outcome evidence defined next.

\subsection{Support-aware deployment and reachability audit}

Let \((\Delta\theta_0,\Delta\omega_{\theta0},\Delta\bar\Omega)\) be the calibrated
nominal effect of the current ground action. Each candidate action is checked
at the takeoff state it would produce, using the mean-wheel approximation for
the budget it would leave:

\(\mathrm{certified}(c)\Longleftrightarrow
  (\hat\theta_0^c,\hat\omega_{\theta0}^c;
  B_\uparrow^c,B_\downarrow^c)\in\mathcal F_\theta\),

\noindent where \(c\) denotes the candidate configuration. An action thus
enters the test twice, moving the queried state and resizing the budget
that bounds the correction, so a favorable pitch shift may still consume
the directional budget that the flight law needs. Because the exact-pair
certificate is sufficient but conservative, DART-S retains it as a
diagnostic audit: a failed query is recorded as uncertified, not unsafe.

For the operational audit, we define an interval terminal set with bounded
landing pitch and pitch rate. Back-propagating this rectangle with the
action-conditioned directional wheel budget gives
\(\mathcal F_\theta^{\mathrm{int}}\). Operational screening requires the
predicted natural state to lie outside this set and the predicted preset state
to lie inside it. A zero-width terminal set gives the stricter exact-pair
\(\mathcal F_\theta\), which remains diagnostic. This interval screen does not
replace local outcome qualification.

This separation is necessary because the measured landing benefit is
non-monotonic in approach speed: a certificate bounds what the airborne
law can still correct, but it does not rank outcomes inside the feasible
region, and simplified takeoff-state overlays reproduce only part of the
measured response map. The speed-map evaluation quantifies both limits
and motivates the outcome-based deployment rule.

Accordingly, operational selection combines two local empirical objects. The
CalMap predicts the takeoff-state shift and reports its interpolation
support; a system-outcome table stores paired DART-S-versus-DART evidence
under the post-touchdown attitude criterion. A preset is deployment-eligible
only where that local outcome evidence is available and the operational
interval transition holds. The exact-pair test is recorded as a diagnostic
reachability audit.

At query time, the selector evaluates the four corners of declared ramp-angle
and speed error bounds. Every checked corner must select the same discrete
action, CalMap must provide support within that ramp-profile class, and the independent
task-outcome evidence must fall within its calibrated radius; otherwise the
selector abstains from preconditioning and returns control to DART. This
four-corner consistency audit is a conservative implementation heuristic,
not a proof that every interior point produces the same decision. Timing
entries use a tighter calibrated radius \(r_\tau\). The resulting branches are
preset, DART-only fallback, and abort by the upstream no-go gate. The
implementation records the evidence source, CalMap support, uncertainty
decision, and nominal/realized interval and exact-pair audits for every
request. After any flight-policy change, we jointly recalibrate outcome
evidence and the action map before further deployment.

To exclude presets in the identified high-speed adverse-response regime, we
augment the empirical selector with a mechanism-based safeguard. On a convex lip, the tire
normal-acceleration margin \(n(v)=g\cos\alpha-v^2/R\) collapses near
\(v^* = \sqrt{gR\cos\alpha}\); a nose-up preset then stretches the lift-off window and
injects an adverse pitch-rate shift. The gate
refuses the preset whenever \(n(v)\le n_{\min}\). A tighter-lip validation
reveals a compensating wheel-spin channel, showing that the adverse
mechanism is geometry-dependent; because the study was
not designed as an equivalence test, the threshold is a calibrated,
ramp-profile-specific safeguard rather than a universal rule, and it
remains consistent with the landing benefit observed across held-out
operating conditions.

Because the stroke is slow relative to the approach, we must complete it
\emph{before the ramp fillet}, where loading would corrupt release; refusal
also needs braking runway. We therefore latch
the gate at a distance
\(d_{\mathrm{latch}}=\max(vt_{\mathrm{act}},
v^2/(2a_{\mathrm{brake,stop}})+d_{\mathrm{margin}})\) before the lip, where
\(v\) is the current approach speed, \(t_{\mathrm{act}}\) is the activation
window, the full-stop decelerating specific force
\(a_{\mathrm{brake,stop}}\) is distinct from DART's shaping deceleration
\(a_{\mathrm{brake,shape}}\), and \(d_{\mathrm{margin}}\) is a stopping margin.

\subsection{Sequential cross-phase allocation}

DART-S coordinates the channels sequentially rather than blending them
simultaneously. Three actuators contribute over two timescales:
suspension presets (slow, ground phase, no direct airborne torque),
fore/aft torque distribution (fast, ground phase, weak authority, kept for
traction), and airborne torque-reaction
(fast, flight phase, protected by an operational spin guard). DART's pitch-axis rate-tracking law
(throttle nose-up, brake nose-down) remains the airborne inner loop and
tracks the terrain-matched landing pitch. This organization requires no fast
feedback coupling loop.
The slow preset and optional lip-relative release pulse map the incoming
approach and selected schedule to a new takeoff state and budget; the
airborne controller then inherits that condition and spends the remaining
budget. Ground preconditioning expends no momentum in flight but changes the
initial wheel-spin distribution; revised audits therefore use measured
per-wheel takeoff speed.

\begin{figure*}[t]\centering
\includegraphics[width=\textwidth]{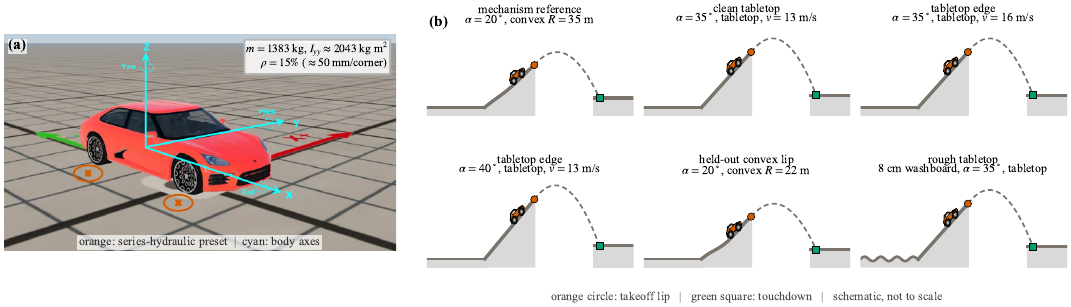}
\caption{DART-S platform and test conditions. (a) DART-S full-scale 4WIDS platform; orange overlays identify the series-hydraulic suspension elements and cyan arrows show the body frame. (b) Representative geometry and disturbance classes: calibration, boundary, and washboard tests use tabletop-fillet ramps; convex $R=35/22$ m exit lips are separate mechanism and held-out conditions. Terrain profiles are schematic and not to scale; orange circles and green squares denote takeoff and touchdown.}\label{fig:darts_platform_scenarios}
\end{figure*}

The current runtime requires coarse inputs: a ramp-angle estimate
at the calibration granularity, onboard approach speed, a lip-curvature
category, a roughness flag, and associated uncertainty bounds. Each decision
requires one CalMap query, one system-evidence lookup, one operational capture
query, and the closed-form margin \(n(v)\); exact-pair certificates, stroke
traces, and statistical audits remain offline diagnostics. Queries outside
local geometry support or with inconsistent input-uncertainty corners are
routed to the audited DART baseline. The selector
perturbation audit specifies the corresponding input-accuracy requirement for
deployment decisions.

\section{Experimental Setup}

All experiments were run in BeamNG.tech build 0.38.3.0 from the v0.38 release
\cite{r29} at full vehicle scale on the same custom SBR4-based 4WIDS platform
as DART; the exact build is retained in run provenance. Vehicle mass is 1383~kg,
\(I_{yy}\approx2043\,\mathrm{kg\,m^2}\), wheel inertia 1.2~kg\(\cdot\)m\(^2\), and radius 0.36~m. Its motor
map has a nominal per-wheel plateau of 350~Nm and a trace-derived effective
per-wheel airborne bound \(\tau_{\max}\approx1200\)~Nm, yielding
\(a_\theta=4\tau_{\max}/I_{yy}\approx2.35\)~rad/s\(^2\) (135$^\circ$/s\(^2\)). For certification,
we use the sustained-effective wheel-speed envelope
\([\Omega_{\min},\Omega_{\max}]=[-125.7,125.7]\) rad/s, distinct from the
200~rad/s operational spin guard and the approximately 221~rad/s drivetrain
hard limit. All configurations share the same per-corner
series-hydraulic hardware. The default per-corner reference-length command
magnitude is approximately 50~mm (\(\rho=15\%\) of the 333~mm neutral
hydro-beam length); the stroke sweep spans approximately 33/50/67~mm
(\(\rho=10/15/20\%\)). At a slew rate of 2 \(s^{-1}\), the full command completes in
$\approx$0.5~s. At zero command, the elements match the stock spring reference
length, so the comparison isolates the preset rather than a hardware change.
Robustness variants span mass $\pm$20\% and a forward CoG shift. Gate latching uses
\(t_{\mathrm{act}}=2.0\)~s, \(a_{\mathrm{brake,stop}}\approx6\)~m/s\(^2\) (versus
DART's \(a_{\mathrm{brake,shape}}=4\)~m/s\(^2\)), and \(d_{\mathrm{margin}}=2\)~m.
Fig.~2(a) shows the platform and actuator layout \mbox{used throughout the study}.

The study covers two classes of ramp profiles. Tabletop-fillet ramps have a
concave entry fillet, a straight face, and an abrupt tabletop termination
without a convex exit lip; they support calibration, boundary, and
rough-runup evaluations. Convex exit-lip kickers with \(R=35\) and 22~m are
reserved for mechanism and held-out-lip studies. Both classes use a 3~m
nominal rise and a flat, gap-style landing.
A proportional approach controller regulates entry speed between 9 and
18~m/s; dedicated boundary and mechanism evaluations extend the range to
40$^\circ$ and 19.5~m/s. At the 35$^\circ$/13~m/s operating point, parameter screening varies preset scale over
\{0.5, 0.75, 1.0\}, ramp rise over \{3, 5, 7\} m, and pre-lip zero-throttle
distance over \{0, 1, 2\} m. The 1 and 2~m cuts are evaluated at \(N=30\),
whereas all other parameter combinations contain \(N=10\) trials per condition.
Figure 2(b) summarizes the ramp profiles and disturbances.

Timing calibration first sweeps
\(t_{\mathrm{lead}}\in\{0.05,0.15,0.25,0.35,0.45,0.55\}\) s at \(N=10\),
then confirms the selected best and worst schedules against neutral and
static presets at \(N=30\). The pulse duration is
\(T_{\mathrm{pulse}}=0.12\)~s and the dimensionless release fraction is
\(\gamma_{\mathrm{release}}=0.50\). At the timing-conditioned operating point,
the confirmed lead time is \(t_{\mathrm{lead}}=0.35\)~s and the dimensionless
normalized support radius is \(r_\tau=0.25\); the command must be restored before
liftoff.
Four-corner stroke traces define a diagnostic pitch-mode phase at 0.2~m
before the lip. This phase is retained for offline mechanism analysis,
whereas the online selector uses the calibrated timing action.

To isolate suspension effects near the operating boundary, we apply the same
symmetric closed-loop approach-speed controller for DART and DART-S. The
common target offset is 0~m/s at the 35$^\circ$/15 and
40$^\circ$/13 conditions and -0.6~m/s at 35$^\circ$/16. DART-S uses no lip throttle cut in
this comparison. We examine the 1~m cut separately as an impact-management
variant while retaining a common no-cut launch profile across all boundary
comparisons.

We instantiate operational interval reachability with terminal half-widths of
35$^\circ$ in pitch and 25$^\circ$/s in pitch rate. Only the directional budget uses the
125.7~rad/s sustained-effective envelope; this value is not a per-wheel
membership box. We constrain online commands at 200~rad/s and use realized
per-wheel trajectories to verify compliance with the 221.2~rad/s drivetrain
limit throughout each run.

The simulator advances deterministically at
\(\Delta t=0.01\,\mathrm{s}\) with a control update at 100~Hz. All analyses retain only records
that satisfy the ballistic time-base check \(\Delta t_{\mathrm{eff}}=\Delta v_z/(gN_{\mathrm{steps}})\). Landing metrics are
latched at the first wheel contact after apex; a fixed-height latch would
be biased since presets change ride height by $\pm$8~cm. Peak deceleration is
the single-step derivative of the mass-aggregated vertical velocity: a
band-limited 10~ms average (-3~dB at 44~Hz) between SAE J211 classes
CFC 10 and CFC 60. Because SAE J211 requires
\(f_s \ge 10\,\mathrm{CFC}\), CFC 10 is the highest compliant class for the
100~Hz state stream; we therefore re-verify load ordering under CFC 10
throughout the study.

We evaluate neutral and preset-only baselines, timing and torque-bias
variants, guarded and unguarded DART/DART-S, landing variants, and the
closed-loop selector.
Both control channels act in the pitch plane by construction: the
transverse wheel spin axes confine momentum exchange to pitch torques,
and the presets command front--rear asymmetry. The experiments therefore
evaluate pitch-plane actuation while retaining roll and yaw in the outcome
bounds.

We assess performance through two complementary endpoints. Actuator, shaping,
ablation, and selector evaluations use a \emph{first-contact criterion}: successful
liftoff followed by
first wheel contact with \textbar pitch\textbar{} $\le$ 35$^\circ$ and \textbar roll\textbar{} $\le$ 30$^\circ$. System-level
evaluations use the stricter post-touchdown criterion, which requires the same
bounds over a 2~s window and treats \textbar pitch\textbar{} \textgreater{} 85$^\circ$ or \textbar roll\textbar{} \textgreater{} 80$^\circ$ as tumble
during that interval.

We randomize controller order within each paired repetition and restart BeamNG
between sessions to limit ordering and simulator-history effects. Each
condition is evaluated across eight independent BeamNG sessions with three
paired repetitions per session. Session-level contrasts define the
inferential unit, while within-session repetitions quantify repeatability.
Primary inference combines equal-weight session effects, a two-level cluster
bootstrap, exact session-level sign-flip tests, leave-one-session-out direction
checks, and Holm correction within each claim family. Confirmatory evidence
must also satisfy prespecified speed-matching, time-integrity,
post-touchdown-window, and hard-limit criteria. We fixed the protocol and
analysis before data collection and retain incomplete trajectories as endpoint
failures. Mechanism interpretation draws on repeated-run studies, whereas
confirmatory inference relies exclusively on independent simulator sessions.

\section{Results}\label{sec:results}

We organize the results around the evidence chain that supports and bounds
DART-S. We first demonstrate system-level boundary performance and then
examine how suspension reshapes the takeoff state and wheel-momentum budget.
We next identify the timing-dependent response and evaluate local selection
before using ablations and profile-specific tests to establish robustness and
scope.

\subsection{Boundary performance of the composed controller}

As shown in Fig.~5(a), DART-S satisfies the post-touchdown attitude criterion
in 24/24 runs at 40$^\circ$/13~m/s, whereas DART satisfies it in 0/24. We observe a
positive effect in every session; the session-equal effect is +1.00 with
cluster interval {[}+1.00,+1.00{]} and boundary-family Holm-adjusted sign-flip
\(p=0.0234\).
Speed matching, leave-one-session-out direction, complete-window, time-base,
and hard-limit gates all pass.

We also observe similarly large attitude-criterion gains at 35$^\circ$/15 and
35$^\circ$/16 (23/24 and 24/24 additional successes over DART). Because entry-speed
dispersion is greater at these conditions, the speed-matched 40$^\circ$/13 result
provides the confirmatory comparison, while the other two characterize
response breadth. The 200~rad/s guard keeps all 600 trajectories within the
221.2~rad/s drivetrain limit.

\subsection{Suspension-induced state and budget shaping}

We first verify that the hydraulic elements deliver the
authority assumed by the state-shift map: full lift and drop move ride
height by +8.3 and -6.3~cm, and the front-up/rear-down preset shifts static
pitch by +3.5$^\circ$/-3.6$^\circ$. Individual corners respond independently, and
damper-mode switching changes the rebound oscillation. The stroke-ratio
sweep shows monotonic takeoff-pitch authority.

Figure 3 plots action-minus-neutral landing-pitch differences. The takeoff
angle, takeoff rate, entry speed, and flight time discussed below are paired
telemetry from the same trials. We use this cross-geometry comparison to test
whether suspension stroke produces a repeatable takeoff-state shift rather
than merely changing approach speed or energy, and compare it with a
same-hardware torque-bias alternative across five ramp angles at 13~m/s.

\begin{figure}[t]\centering
\includegraphics[width=\columnwidth]{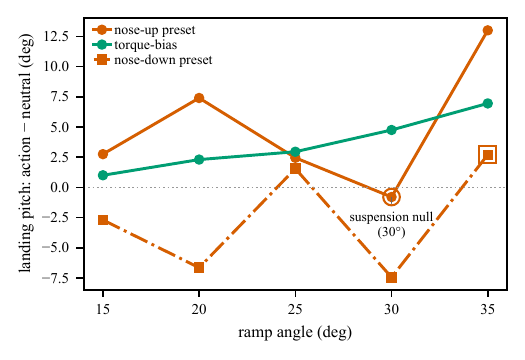}
\caption{Action-minus-neutral landing-pitch difference vs.\ ramp angle at $v=13$ m/s ($N\ge30$ per condition) for the suspension presets and torque-bias baseline. Open markers: not significant ($p\ge0.05$).}\label{fig:darts_e2}
\end{figure}

Across the five geometries, we observe a stable nose-up pitch bias
of +2.2\ldots+3.4$^\circ$ with takeoff-rate shifts of 0\ldots+7~$^\circ$/s, while entry speed and
flight time remain unchanged to within
1\%. A ballistic estimate,
\(\Delta_{\mathrm{land}}\approx\Delta\theta_0+\Delta\omega_{\theta0}T\), captures the sign and ordering of the landing gains
but overpredicts their magnitude by a near-constant 1.5--3.3$^\circ$ because of
in-flight relaxation from wheel spin-down and the measured aerodynamic
pitching moment; the estimate therefore serves as an upper bound. At the
plotted 30$^\circ$ points, this relaxation cancels the rate-free
static shift (-0.8$^\circ$, \(p=0.054\)), whereas the torque-bias configuration
retains a +4.8$^\circ$ landing improvement; the contrast motivates per-condition
gating of the preset.

Torque bias has negligible, opposite-sign takeoff-angle authority
(-0.2\ldots-0.7$^\circ$). A separate matched-\(v_0\) throttle-cut counterfactual, not
plotted in Fig.~3, reduces its 30$^\circ$/35$^\circ$ landing gains from +4.8$^\circ$/+7.0$^\circ$ to
+1.6$^\circ$/+2.0$^\circ$, identifying a
launch-profile spin effect rather than a schedulable attitude channel.
Mirroring the suspension command confirms directionality: the plotted 20$^\circ$
nose-down point shows a 6.7$^\circ$ landing-pitch degradation; companion outcome
records show the pass rate falling from 1.0 to 0.067.

\subsection{Local selection and non-monotonic response}

\begin{figure}[t]\centering
\includegraphics[width=\columnwidth]{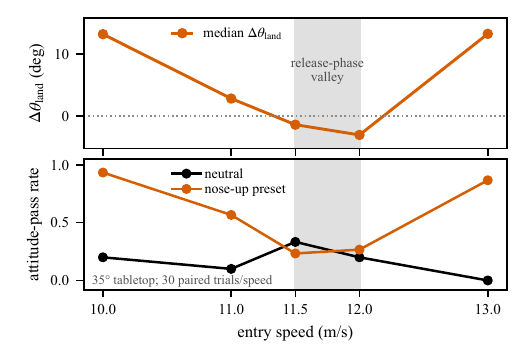}
\caption{Speed-dependent response on the 35$^\circ$ tabletop fillet (30 pairs per speed). Positive static-pitch authority persists through the shaded 11.5--12~m/s landing-response reversal, motivating timing-conditioned action selection.}\label{fig:darts_e3}
\end{figure}

A useful preset must improve a condition, not merely have the correct static
sign. As shown in Fig.~4, we evaluate the 35$^\circ$ tabletop fillet at five speeds.
Median preset-minus-neutral landing-pitch changes are
+13.20$^\circ$, +2.85$^\circ$, -1.35$^\circ$, -3.00$^\circ$, and +13.25$^\circ$ at 10, 11, 11.5, 12, and 13~m/s.
The prespecified global valley contrast is -14.93$^\circ$ (bootstrap 95\% CI
-16.46\ldots-13.34$^\circ$, Holm-adjusted one-sided Wilcoxon
\(p=1.86\times10^{-9}\)); the local contrast is -1.84$^\circ$ (95\% CI
-3.52\ldots-0.13$^\circ$, \(p=0.0303\)). All three sessions agree in sign. These data
support an 11.5--12~m/s release-phase valley. Five of 300 runs at 10~m/s
lack complete post-touchdown windows and are counted as failures, keeping the
overall evidence partial. We nevertheless observe a repeated sign reversal
consistent with a release-transient mechanism under this geometry.

Static takeoff-pitch authority remains +2.2\ldots+3.1$^\circ$ throughout, whereas the
release-rate contribution falls from +9.0~$^\circ$/s at 10~m/s to +0.5/0.0~$^\circ$/s at
11.5/12~m/s and then recovers to +6.0~$^\circ$/s; this channel separation, not the
dip alone, is the central insight. Approach speed therefore indexes the synchronization
among suspension rebound, front-axle unloading, and lip departure rather than
the magnitude of static authority. The same final preset can consequently
improve or degrade landing. A ballistic overlay matches 12/16 exploratory
points but misses the valley; a rate-only fit (RMSE 0.136~deg/s) detects the
13 and 14~m/s transition but not displacement feasibility. This rules out a
globally enabled preset and any monotonic speed interpolation for deployment.

We introduce a zero-final-offset timing intervention to isolate the timing effect at
11.5~m/s. Across eight independent BeamNG sessions, the 0.35~s timing
action satisfies the post-touchdown criterion in 23/24 runs versus 0/24 for
the static preset; all session effects are positive (cluster interval
{[}+0.833,+1.000{]}, Holm-adjusted \(p=0.0156\)), and the
speed-match gate passes. The 0.15~s action also satisfies the criterion in
0/24, but its comparison
fails the stricter timing speed-match gate and remains descriptive.
Conditioning identifies the realized takeoff state, rather than the phase
proxy, as the mediator linking timing to landing response at this condition.

Away from the operating boundary, we observe mildly sub-additive continuous
landing-pitch gains: at 35$^\circ$ the combined +22.0$^\circ$ gain is 14\% below the
single-channel sum, with the same ordering at 20$^\circ$. The preset removes part
of the error before liftoff, leaving less for the airborne controller.
This saturation of a continuous error metric does not predict the binary
boundary outcome, where crossing a failure threshold yields the positive
interaction reported below.

By varying the stroke ratio and command endpoint, we localize the gain to
takeoff-state shaping: increasing the per-corner command from approximately
33 to 67~mm (\(\rho=10\%\) to \(20\%\)) raises
\(\Delta\theta_0\) monotonically
(+2.5$^\circ$/+3.3$^\circ$/+4.0$^\circ$) and increases \(\Delta_{\mathrm{land}}\)
(+2.8$^\circ$/+7.4$^\circ$/+8.2$^\circ$), linking commanded suspension authority to the eventual
landing response rather than to an incidental launch-speed change. Under
mass $\pm$20\% and a forward CoG shift, \(\Delta_{\mathrm{land}}\) remains +6.5\ldots+7.5$^\circ$ around
the +7.4$^\circ$ nominal result. Conversely, landing-end suspension variants change
pitch by at most 0.6$^\circ$ and peak deceleration by at most $\pm$4.8\%. Table I
therefore localizes the useful authority to pre-liftoff shaping and shows
that it persists under moderate changes in vehicle inertia and mass
distribution.

As summarized in Table I, we verify pass-through, refusal, preparation, and
the requirement that preset stroke complete before fillet loading. These
tests isolate decision execution and stroke timing, providing the basis for
the subsequent reachability and branch-fidelity analysis reported below.

\begin{table*}[t]\centering\footnotesize
\caption{Compact validation summary for actuator authority, ablations, robustness, and selector execution. \(\Delta_{\mathrm{land}}\) ($^\circ$) is reported relative to neutral.}\label{tab:e6}
\resizebox{\textwidth}{!}{\begin{tabular}{lll}
\toprule\noalign{}
test
 & protocol
 & measured result and interpretation
 \\
\midrule\noalign{}

static authority & full lift/drop; front-up/rear-down & ride height changes by +8.3 and -6.3~cm; static pitch is +3.5$^\circ$/-3.6$^\circ$ \\
stroke scaling & 33/50/67~mm (10/15/20\%), \(N=30\) & \(\Delta\theta_0\) = +2.5$^\circ$/+3.3$^\circ$/+4.0$^\circ$; monotonic takeoff authority \\
takeoff vs both ends & 20$^\circ$ and 35$^\circ$ & +7.0$^\circ$ versus +7.4$^\circ$ at 20$^\circ$; +8.6$^\circ$ versus +13.0$^\circ$ at 35$^\circ$ \\
platform robustness & mass $\pm$20\%; CoG-forward, \(N=30\) & \(\Delta_{\mathrm{land}}\) between +6.5$^\circ$ and +7.5$^\circ$ (all \(p<10^{-4}\)), versus +7.4$^\circ$ nominal \\
landing-gain stroke scaling & 33/50/67~mm (10/15/20\%) & \(\Delta_{\mathrm{land}}\) = +2.8$^\circ$/+7.4$^\circ$/+8.2$^\circ$ \\
landing-end variants & \(N=30\) & pitch change at most 0.6$^\circ$; peak-deceleration change within $\pm$4.8\% (not an equivalence test) \\
selector execution & pass-through/refuse/prepare & 30/30 decisions; 30/30 stopped, median margin 7.3~m; prepare 28/30 vs independent reference 0/115 \\
selector timing & lead 2.0 vs 0.8~s & prepare success 28/30 vs 21/30; preset must finish before the fillet \\
\bottomrule
\end{tabular}}\end{table*}

Across 3465 queries at 21 calibrated conditions, decisions are invariant
within $\pm$0.5$^\circ$/$\pm$0.2~m/s and through $\pm$2.5$^\circ$ angle error. All 66 harmful flips
occur at the timed condition for speed error $\ge$0.25~m/s, so larger uncertainty
disables that entry. The curvature safeguard requires \textgreater0.6~m/s speed
underestimate to admit the nearest measured adverse condition.

We observe that the prespecified selector executes all 48 positive
interpolation branches, all 24 negative branches, and all 24
uncertainty-boundary abstentions as specified. The realized interval audit
accepts 42/48 positive selected actions. State-only reachability always
retains the natural state, whereas
budget-only, joint, and outcome-only policies choose identical actions in
the tested conditions. Thus the observed reachability distinction is budget-driven,
but it adds no action change beyond the locally qualified outcome table.

We also compare DART, selector fallback, and forced preset on the new
washboard evaluation; each satisfies the attitude criterion in 24/24 runs.
The prespecified selector-versus-forced contrast is therefore zero; the
earlier protective washboard claim is not retained. Two held-out interpolation
conditions show large descriptive effects (24/24 versus 0/24 and 24/24 versus
2/24). Because their entry-speed dispersion exceeds the confirmatory
tolerance, we treat these results as supporting trends for future
speed-matched validation.

\subsection{Ablations, robustness, and scope}

\begin{figure}[t]\centering
\includegraphics[width=\columnwidth]{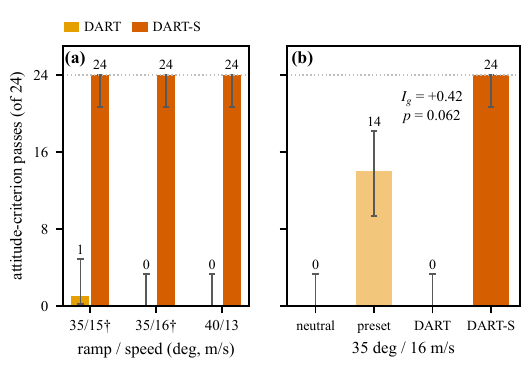}
\caption{Boundary confirmation across independent BeamNG sessions (post-touchdown attitude-criterion successes out of 24; Wilson 95\% intervals). (a) At 40$^\circ$/13~m/s, DART-S attains 24/24 versus 0/24 for DART. Daggered conditions have large descriptive effects but fail their prespecified speed-match gates. (b) The 35$^\circ$/16~m/s factorial yields a positive interaction estimate, but its session-level sign-flip test is not significant ($p=0.0625$).}\label{fig:darts_e9}
\end{figure}

As shown in Fig.~5(b), the 35$^\circ$/16~m/s factorial yields criterion-success
counts of 0/24 for neutral and DART, 14/24 for preset-only, and 24/24 for
outcome-only, budget-only, and joint DART-S. The session-equal factorial
interaction is +0.417 with cluster interval {[}+0.125,+0.708{]}, but the exact
session-level test gives \(p=0.0625\). This ordering is consistent with
complementary ground and airborne contributions; a larger session set is
needed to resolve the interaction magnitude. Historical guard
ablations show that guarded trajectories stay below the drivetrain limit,
whereas unguarded trials frequently cross it during airborne correction,
directly exposing how the guard preserves wheel-speed headroom throughout the
remaining airborne control horizon of each jump.

We next compare the nominal mean-wheel approximation with realized per-wheel
means and observe a median absolute error of 15.85~rad/s and a maximum error
of 152.28~rad/s.
Every strategy run has at least one realized wheel beyond the 125.7~rad/s
certificate envelope, 60/240 exceed 200~rad/s on at least one wheel, and none
exceeds the 221.2~rad/s hard limit. Reachability is therefore reported as an
operational screen derived from the nominal state/budget model, with realized
signed per-wheel traces providing the authority record. Landing-load shaping
also remains open:
historical guarded CFC-10 medians are approximately 33 g for DART and 43 g
for DART-S at this boundary condition.

\begin{figure*}[!t]\centering
\includegraphics[width=\textwidth]{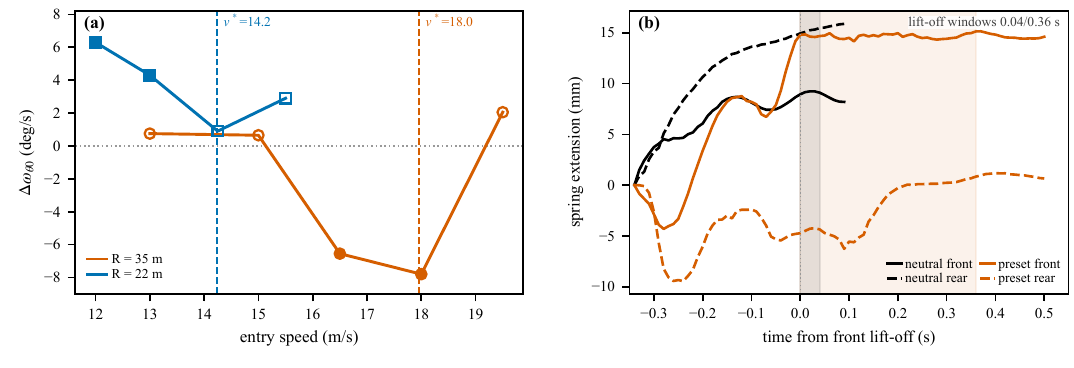}
\caption{Mechanism measurements. (a) Paired takeoff pitch-rate difference $\Delta\omega_{\theta0}$ vs.\ entry speed on the 20$^\circ$ kicker (calibration $R=35$~m, held-out $R=22$~m); dashed lines mark $v^*=\sqrt{gR\cos\alpha}$; filled markers: sign-test $p<0.05$. (b) Overlaid axle-stroke traces at 18~m/s ($R=35$~m): the nose-up preset stretches the lift-off window from 0.04 to 0.36~s (shaded).}\label{fig:darts_mech}
\end{figure*}

As shown in Fig.~6(a), we observe a profile-specific high-speed reversal on
DART's 20$^\circ$ reference kicker (\(R=35\)~m): the nose-up preset retains
+2.5\ldots+4.5$^\circ$ of takeoff-pitch authority (\(p<10^{-3}\)), yet median landing
error worsens by +3.4$^\circ$ at 18~m/s \((N=30,\ p<10^{-4})\) as the takeoff-rate
shift reverses. The reversal is therefore governed by pitch-rate and
wheel-momentum dynamics despite the retained static pitch bias.

For an ideal convex lip of radius R, the quasi-static normal-load margin per
unit mass is \(n(v)=g\cos\alpha-v^2/R\) and crosses zero at
\(v^* = \sqrt{gR\cos\alpha}\) = 17.96~m/s for this geometry. On the paired speed ladder
(Fig.~6(a)), no pitch-rate difference is detected at 13 or 15~m/s
(n $\ge$ 2.8~m/s\(^2\)); adverse shifts of -6.6 and -7.8~$^\circ$/s emerge at 16.5 and
18~m/s (sign-test \(p\le0.002\)) as \(n(v)\) approaches $\approx$1.4~m/s\(^2\), and no significant
shift is detected past \(v^*\) at 19.5~m/s.

We use the stroke traces in Fig.~6(b) to expose the mechanism at 18~m/s.
With neutral suspension both axles leave the lip nearly together
(front-to-rear lift-off window 0.03--0.04~s). The nose-up preset holds the
front corners extended: the depleted normal margin releases the front axle
early while the compressed rear stays loaded, stretching the window to a
median 0.36~s, during which the vehicle pivots nose-down about the rear
contact. The traced rate penalty
(-9.1~$^\circ$/s, $\approx$ -28~$^\circ$/s per second of window) is consistent in magnitude with
the ladder response.

We then use a held-out validation to test whether \(n(v)\) transfers across
geometries. Moving the lip to \(R=22\)~m shifts \(v^*\) to 14.2~m/s, yet the held-out ladder
shows no adverse band (Fig.~6(a)); \(\Delta\omega_{\theta0}\) is \emph{positive} at 12 and
13~m/s (+6.3 and +4.3~$^\circ$/s, \(p\le0.04\)), with no significant shift detected
past \(v^*\). A 2$\times$2 radius-by-arc-length comparison at n $\approx$ 0 (3.05~m
versus 4.89~m of arc) shows that neither variable alone is sufficient: the adverse shift
persists on the short \(R=35\)~m arc (-10.0~$^\circ$/s, \(p=0.002\)), reappears
on the long \(R=22\)~m arc (-6.4~$^\circ$/s, \(p=0.021\)), and the short \(R=22\)~m lip
remains benign.

Finally, we synthesize traces across all four operating conditions to resolve the
conflict into two channels with one root. Across every condition, the preset stretches the lift-off window
(+0.19\ldots+0.33~s), and strong nose-down pitch acceleration persists inside
that window (-50\ldots-121~$^\circ$/s\(^2\)); this is the adverse channel. But the
early-unloaded
front wheels are also spun up by the drivetrain during lip transit, and
their nose-up reaction torque is largest where the neutral configuration
stays gripped while the preset unloads (takeoff wheel-speed difference up
to +95~rad/s on the tight lip, near zero on \(R=35\)~m where both
configurations saturate); this is the compensating channel. Competition
between the two channels sets the net response, with near cancellation on the
short, tight lip. A calibrated threshold,
\(n(v)\le n_{\min}\) (\(n_{\min}=2\,\mathrm{m/s^2}\) on the \(R=35\)~m ladder),
intercepts every measured adverse condition. Where it activates under a
canceling spin channel, no significant gain is forfeited, although this study
does not establish zero intervention cost. Accordingly, \(n(v)\) serves as a
ramp-profile-specific indicator of contact-timing reversal and supplies a
conservative exclusion boundary for deployment.

\section{Conclusion}

DART-S turns active suspension into a pre-liftoff state-and-budget actuator:
the ground phase reshapes takeoff pitch, pitch rate, and wheel headroom before
the guarded DART law spends the remaining authority in flight. A support-aware
selector combines local calibration, outcome evidence, and operational
interval-reachability screening; exact-pair reachability provides the
diagnostic audit.

Confirmation across independent BeamNG sessions establishes two local
results: DART-S
satisfies the post-touchdown criterion in 24/24 runs versus 0/24 for DART at
40$^\circ$/13~m/s, and the 0.35~s timing action does so in 23/24 versus 0/24 for the
static preset. Prespecified positive, negative,
and boundary branches execute as specified, and the guard keeps every
trajectory within the drivetrain limit. The speed-matched boundary and timing
comparisons define the confirmatory scope; additional conditions characterize
response breadth and identify priorities for higher-powered session
replication. On the washboard, DART, fallback, and forced preset each satisfy
the criterion in 24/24 runs.

This study establishes reachability-audited authority within the validated
simulation envelope. Future hardware validation will extend this evidence
through loaded-hydraulic actuator measurements, scaled-vehicle jumps,
structural loads, and roll-axis evaluation.

\section*{Acknowledgments}
During the preparation and technical implementation of this work, the
authors utilized a combination of generative artificial intelligence (AI)
systems in an integrated workflow. Specifically, Auto/Composer 2.5 (Cursor),
GLM 5.3, DeepSeek V3/V4, and Kimi K3 were used interactively to assist in developing,
debugging, and optimizing the algorithmic source code and data-processing
pipelines discussed in Section~\ref{sec:method}. These same systems
(GLM 5.3, DeepSeek V3/V4 and Kimi K3) were also co-employed to generate the analytical
visualization scripts for creating the experimental data charts in
Section~\ref{sec:results}. During the manuscript revision phase, GLM 5.3, DeepSeek V3/V4
and Kimi K3 refined English prose, enhanced language fluency, and polished
the text across all sections. In accordance with IEEE guidelines, all
AI-assisted engineering code, data visualizations, and textual outputs were
thoroughly reviewed, logically validated, and cross-checked against raw
experimental results by the human authors, who retain full academic and
ethical responsibility for the content and integrity of this publication.

\balance

\end{document}